\pdfoutput=1
\documentclass[letterpaper,english]{article} 
\usepackage{lmodern}
\usepackage[T1]{fontenc}
\usepackage[latin9]{luainputenc}
\usepackage{geometry}
\usepackage{babel}
\usepackage{float}
\usepackage{amsmath}
\usepackage{amsthm}
\usepackage{amssymb}
\usepackage{graphicx}

\makeatletter

\providecommand{\tabularnewline}{\\}
\floatstyle{ruled}
\newfloat{algorithm}{tbp}{loa}
\providecommand{\algorithmname}{Algorithm}
\floatname{algorithm}{\protect\algorithmname}

\theoremstyle{plain}
\newtheorem{thm}{\protect\theoremname}[section]
  \theoremstyle{plain}
  \newtheorem{assumption}[thm]{\protect\assumptionname}
  \theoremstyle{plain}
  \newtheorem{lem}[thm]{\protect\lemmaname}
  \theoremstyle{plain}
  \newtheorem{cor}[thm]{\protect\corollaryname}

\usepackage{proceed2e} 
\usepackage{times}

\makeatother

  \providecommand{\assumptionname}{Assumption}
  \providecommand{\corollaryname}{Corollary}
  \providecommand{\lemmaname}{Lemma}
\providecommand{\theoremname}{Theorem}

\title{Randomized Smoothing SVRG for Large-scale Nonsmooth Convex Optimization}
\author{} 

\author{ {\bf Wenjie Huang} \\
	Dept. of Industrial Systems Engineering and Management\\
	National University of Singapore\\
	wenjie\_huang@u.nus.edu \\
}

\begin{document}

\maketitle
\begin{abstract}
In this paper, we consider the problem of minimizing the average of
a large number of nonsmooth and convex functions. Such problems often
arise in typical machine learning problems as empirical risk minimization,
but are computationally very challenging. We develop and analyze a
new algorithm that achieves robust linear convergence rate, and both
its time complexity and gradient complexity are superior than state-of-art
nonsmooth algorithms and subgradient-based schemes. Besides, our algorithm
works without any extra error bound conditions on the objective function
as well as the common strongly-convex condition. We show that our
algorithm has wide applications in optimization and machine learning
problems, and demonstrate experimentally that it performs
well on a large-scale ranking problem. 
\end{abstract}

\section{Introduction}

In this paper, we develop and analyze stochastic variance reduction
algorithm with randomized smoothing techniques for solving the following
class of large-scale nonsmooth optimization problems. Problems of
this form often arise in machine learning and statistics, as the regularized
empirical risk minimization for a certain class of statistical learning
problems.

We consider the problem of minimizing the sum of two convex functions:\index{}
\begin{equation}
\min_{x\in\mathbb{R}^{d}}\quad\left\{ P(x)\triangleq F(x)+R(x)\right\} ,\label{Main problem}
\end{equation}
where $F(x)$ is the average of many component functions $f_{i}(x)$,
i.e.,
\[
F(x)=\frac{1}{N}\sum_{i=1}^{N}f_{i}(x),
\]
where each component function $f_{i}$ is convex and proper function,
but not necessarily smooth, and $R:\,\mathbb{R}^{d}\rightarrow\mathbb{R}$
is a known regularizing function. We assume that $R$ is closed and
convex, but we also allow for nondifferentiability so that the framework
includes the $l_{1}$-norm and related regularizers. The problem \ref{Main problem}
is challenging for three reasons. First, the $f_{i}$ may be nonsmooth.
Second, in the case where the number of components $N$ is very large,
it can be advantageous to use incremental methods (such as stochastic
subgradient descent) that operate on a single component $f_{i}$ at
each iteration, rather than on the entire cost function. This will
cause slow convergence and high overall computation complexity. Third,
$f_{i}$ may not simply be strongly convex, which is particularly
true for Lasso and $l_{1}$-Regularized Logistics Regression. 

The convergence rate and computational complexity analysis of subgradient
descent and stochastic subgradient descent in non-strongly convex
optimization has widely been studied in literature. In (\cite{Shamir2013}),
optimal averaging schemes are applied to stochastic gradient descent
to achieve $O(\log(T)/\sqrt{T})$ for nonsmooth convex objective functions,
and $O(\log(T)/T)$ in the nonsmooth strongly-convex case, after $T$
iterations. In (\cite{Bach2013}), the convergence rate $O(1/T)$ is
achieved without strongly-convex condition. In (\cite{Shalev-Shwartz2009}),
the authors show that strong convexity and regularization are key
ingredients for the uniform convergence of empirical minimization
problem in general case. In (\cite{Yang2015}), the authors study the
efficiency of a Restarted subgradient method that periodically restarts
the standard subgradient method, and show that it has a lower complexity
than stochastic subgradient descent, and also reduce the dependence
on the initial solution. 

Various error bounds are studied to improve the convergence rate results
for nonsmooth opitmization. The error bound summarizes the relationship
between the distance to the optimal objective value, with respect
to the distance to the set of optimal solutions. Those bounds include:
Polyhedral and Quadratic error bound (see (\cite{Yang2015a,Johnstone2017})), Local growth
rate (see (\cite{Xu2017})), Strong error bound condition under randomized
methods (see (\cite{Nedic2001})), Inverse growth condition (See (\cite{Robinson1999})).
The limitations of imposing error bound condition on model and algorithm
are obvious. First, only a small class of problems conforming
to certain global or local error bound condition have a superior convergence
rate or complexity, such that the results cannot be extended to wider
class of optimization problems. Second, it does not change the problem
and algorithm structure, thus it can not be treated as a direct improvement. 

In (\cite{Maekelae2002}), Bundle method has been investigated in
nonsmooth optimization. The basic idea of Bundle method is in each
iteration to use subgradient information on historical solution estimations
to generate first-order Taylor approximation as lower bound. As the
algorithm evolves, the solution will be improved as the lower bound
becomes tighter. In (\cite{Le2008}), bundle method is applied into
solving large-scale nonsmooth optimization problems in machine learning
i.e., Support vector machine, and it shows to achieve $O(1/T)$ for
general convex problem and linear convergence rate for continuously
differentiable problems. The difficulty of such approaches is that
it requires quite detailed knowledge of the structure of the objective
function in order to have an accurate lower approximation. In (\cite{Mifflin1998,Shen2013}), Moreau-Yosida regularization, Bundle method, and quasi-Newton method
are combined to achieve superlinear convergence for nondifferentiable
convex optimization problems. In (\cite{Nesterov2005}), the authors
propose a new approach for constructing efficient schemes for nonsmooth
convex optimization. The smoothing technique is based on convex conjugate
function with regularization in dual space, and the dual averaging
acceleration manage to improve the traditional bounds on the number
of iterations of the gradient schemes. However, the current smoothing
techniques have not taken the advantage of finite sum structure. Bundle
method as well as dual averaging method require large gradient complexity
in each iteration, in particular, subgradients of each component for
all historical solutions. Thus new smoothing method is required to largely
reduce the computational complexity of gradient. 

Variance reduction is an important issue in large-scale optimization
with gradient-based approaches. In (\cite{Johnson2013}) and its proximal extension in (\cite{Xiao2014}), stochastic
variance reduced gradient (SVRG) is proposed that reduces the variance
of stochastic gradient descent, and enjoys the same fast convergence
rate as those of stochastic dual coordinate ascent (SDCA) proposed
in (\cite{Shalev-Shwartz2009,Shalev-Shwartz2012}) and stochastic average
gradient (SAG) proposed in (\cite{Schmidt2017}). In (\cite{Gong2014,Allen-Zhu2016}),
SVRG achieves linear convergence without strongly-convexity condition. The standard
SVRG is extended to second-order methods: Stochastic quasi-Newton and L-BFGS in (\cite{Mairal,Lucchi2015,Moritz2016}), . In
(\cite{Vainsencher2015}), the authors study how local smoothness helps
to improve SVRG.

To the best of our knowledge, there is no systematic work on large-scale
nonsmooth optimization problems for which a reduction in the variance
of the stochastic estimate of the subgradient on approximated function
gives an improvement in convergence rates. We use randomized smoothing
technique in (\cite{Duchi2012}), to smooth each component function,
where convolution-based smoothing technique is applied. The intuition
underlying such approaches is that the convolution of two functions
is at least as smooth as the smoothest of the two original functions.
Then a proper random perturbation of variable can transform objective
into a smooth function. Then we apply proximal version of SVRG on the smoothed function,
which takes the advantage of the problem structure, average of deterministic
component functions, to achieve much superior convergence rate and
computational complexity than nonsmooth convex optimization algorithms
in (\cite{Nesterov2005,Mifflin1998,Le2008}) as well as traditional
subgradient-based schemes. Besides, our algorithm works without strongly-convex
requirement on the objective. 

\subsection{Notations and Assumptions}

We presents the notations used throughout this paper. Define the minimizer
of the objective function as $x_{\ast}:=\arg\min_{x\in\mathbb{R}^{d}}\,P(x).$
We define $\mathcal{B}_{p}(x,\,u)=\{y\in\mathbb{R}^{d}|\|x-y\|_{p}\leq u\}$
to be the closed $p$-norm ball of radius $u$ around the point $x$.
Addition of sets $A$ and $B$ is defined as the Minkowski sum in
$\mathbb{R}^{d}$, $A+B=\{x\in\mathbb{R}^{d}|x=y+z,\,y\in A,\,z\in B\}$,
multiplication of a set $A$ by a scalar $\alpha$ is defined to be
$\alpha A:=\{\alpha x|x\in A\}$, and $\textrm{aff}(A)$ denotes the
affine hull of the set $A$. We let $\textrm{supp}\mu:=\{x|\mu(x)\neq0\}$
denote the support of a function or distribution $\mu$. We use $\partial f(x)$
to denote the subdifferential set of the convex function $f$ at point
$x$. Given a norm $\|\cdot\|$, we adopt the shorthand notation $\|\partial f(x)\|=\sup\{\|g\|\,|g\in\partial f(x)\}$.
The dual norm $\|\cdot\|_{\ast}$ associated with a norm $\|\cdot\|$
is given by $\|z\|_{\ast}:=\sup_{\|x\|\leq1}\left\langle z,\,x\right\rangle $.
The gradient of $f$ is $L_{1}$-Lipschitz continuous with respect
to the norm $\|\cdot\|$ over some closed convex set $\mathcal{X}\subseteq\mathbb{R}^{d}$
if
\[
\|\nabla f(x)-\nabla f(y)\|_{\ast}\leq L_{1}\|x-y\|,\quad\forall x,\,y\in\mathcal{X}.
\]
The transpose of matrix $X$ is denoted as $X^{\top}$. Given a random
variable $\xi$ drawn from the distribution $P$, $P$-a.e. $\xi$
is the shorthand for $P$-almost every $\xi$. 

We have the following assumptions throughout this paper. 
\begin{assumption}
\label{Assumption 1.1} (i) The function $R(x)$ is lower semi-continuous
and convex, and its effective domain, $\textrm{dom}(R):=\{x\in\mathbb{R}^{d}|R(x)<+\infty\}$
is closed. 

(ii) The function $F$ is $L_{0}$-Lipschitz with respect to the $l_{2}$-norm
$\|\cdot\|_{2}$ over $\textrm{dom}(R)\subseteq\mathbb{R}^{d}$, that
$|F(x)-F(y)|\leq L_{0}\|x-y\|_{2},\,\forall x,\,y\in\textrm{dom}(R).$
\end{assumption}

\subsection{Our contributions }

In this paper, we develop new method for nonsmooth empirical risk
minimization problems for which reductions in the variance of the
stochastic estimation of the true subgradient, as well as in the variance
of stochastic gradient after randomized smoothing by employing a multistage
scheme, give an significant improvement in computation. Our algorithm
has both advantages in fast convergence rate and low total computational
complexity. 
\begin{enumerate}
\item First, to the best of our knowledge, there is currently no method
in literature that can achieve linear convergence rate for optimization
problem with finite sum of nonsmooth objectives, thus our method can
largely improve the computational efficiency for estimation and optimization
on large-scale data sets for many machine learning problems with nonsmooth
objectives. 
\item Second, our algorithm are developed based on composite objective functions
that can solve more general class of problems. Our algorithm works
even both $F$ and $R$ are non-strongly convex. Actually, same reduced
variance bounds and convergence rate holds for problems with or without
composite functions $R(x)$.
\item Third, our algorithm has total complexity, given the $\epsilon$-optimal
solution, the time complexity of $O(\log(1/\epsilon))$ and the gradient
complexity is $O(N\cdot m\cdot\log(1/\epsilon)+1/\epsilon)$, where
$m$ is the number of samples in the randomized smoothing procedure.
This complexity is far superior than that of other gradient based
approaches and many classical accelerated and smoothing methods in
nonsmooth optimization which only achieve sublinear convergence rate.
In particular, full subgradient descent after randomized smoothing
gives the time complexity $O(1/\epsilon)$, and the gradient complexity
$O(N\cdot m/\epsilon).$ From (\cite{Bach2013}), we can conclude that
the stochastic gradient descent after randomized smoothing gives the
complexity $O(m/\epsilon)$. For both stochastic and normal subgradient
descent without smoothing, the complexity is $O(1/\epsilon^{2})$
in non-strongly-convex case, which is extremely poor. For Bundle method
and Nesterov's smoothing method, the complexity is normally $O(1/\epsilon)$.
The time complexity of SDCA for nonsmooth objectives is $O(1/\epsilon)$
and the gradient complexity is $O((N+1)/\epsilon)$. For SAG, the
time complexity for nonstrongly convex objective is $O(1/\epsilon)$,
but the theoretical analysis of SAG working on nonsmooth objectives
has not been established. 
\end{enumerate}
The remainder of the paper is organized as follows. In Section 2,
we give a full description of our randomized smoothing SVRG (RS-SVRG)
algorithm, and state the main theorems. In Section 3, we discuss some
applications of our method and provide experimental results illustrating
the merits of our approach. Finally, we concludes the paper in section
4, with certain more technical aspects deferred to the supplymental
material. 

\section{Main Results}

We begin by motivating the algorithm studied in this paper, and we
then state our main results on its convergence. 

\subsection{Description of the Algorithm}

The starting point for our approach is a convolution-based smoothing
technique amenable to nonsmooth stochastic optimization problems.
A number of authors have noted that random perturbation of the variable
$x$ can be used to transform $f$ into a smooth function. The intuition
underlying such approaches is that the convolution of two functions
is at least as smooth as the smoothest of the two original functions.
In particular, letting $\mu$ denote the density of a random variable
with respect to Lebesgue measure, consider the smoothed objective
function
\begin{equation}
F_{\mu}(x)=\int_{\mathbb{R}^{d}}F(x+y)\mu(y)dy=\mathbb{E}_{\mu}\left[F(x+Z)\right],\label{Smoothing}
\end{equation}
where $Z$ is a random variable with density $\mu$. Clearly, the
function $F_{\mu}$ is convex when $F$ is convex; moreover, since
$\mu$ is a density with respect to Lebesgue measure, the function
$F_{\mu}$is also guaranteed to be differentiable. 

We analyze minimization procedures that solve the nonsmooth problem
(\ref{Main problem}) by using stochastic gradient samples from the
smoothed function with appropriate choice of smoothing density $\mu$.
The RS-SVRG algorithm is presented as Algorithm 1. Given an initial
solution $x^{\phi}$, our algorithm is divided into $s$ epoches.
The $s$-th epoch consists of randomized smoothing steps for all $N$
component functions (Steps (a),(b), and (c)) and $M_{s}$ stochastic
gradient steps (Steps (e) and (f)), where $M_{s}$ doubles between
every consecutive epoches (see Step (d) in Algorithm 1). This \textquotedblleft doubling\textquotedblright{}
feature follows the $\textrm{SVRG\ensuremath{^{++}}}$ developed in
(\cite{Allen-Zhu2016}), and distinguishes our method from traditional
variance-reduction based methods. Our starting solution $x^{\phi}$
of each epoch is set to be the ending solution $x_{M_{s}}$ of the
previous epoch (see Step (g) in Algorithm 1), rather than the average
of the previous epoch in (\cite{Xiao2014}), and randomly selection
from the solutions in previous epoch in (\cite{Johnson2013}). In Algorithm
1, we define the proximal mapping $\textrm{Pro}x_{R}$ as
\[
\textrm{Pro}x_{R}(y)=\arg\min_{x\in\mathbb{R}^{d}}\left\{ \frac{1}{2}\|x-y\|_{2}^{2}+R(x)\right\} .
\]

\begin{algorithm}
\caption{Randomized Smoothing SVRG}

Input: initial solution $x^{\phi}$, and set $x_{0}=x^{\phi}$, step-sizes
$\{\gamma_{s}\}_{s\geq0}$, inner iterations $M$. Set $t=0$.
\begin{enumerate}
\item For $s\geq1$
\begin{enumerate}
\item Set $\tilde{x}=\tilde{x}_{s-1}$.
\item Draw random variable $\{Z_{j,s}\}_{j=1}^{m}$ i.i.d according to the
distribution $\mu$. 
\item Compute the subgradient $g_{i,j,s}(\tilde{x})\in\partial f_{i}(\tilde{x}+a_{s}Z_{j,s})$,
and compute $\tilde{g}_{i}(\tilde{x})=\frac{1}{m}\sum_{j=1}^{m}g_{i,j,s}$,
and $\tilde{g}=\frac{1}{N}\sum_{i=1}^{N}g_{i}(\tilde{x})$.
\item $M_{s}\leftarrow2^{s}\cdot M$
\item For $t=1,\ldots,\,M_{s}$
\begin{enumerate}
\item Randomly pick $\mathcal{I}_{t}\in\left\{ 1,\ldots,\,N\right\} $,
and set $v_{t}=\tilde{g}_{\mathcal{I}_{t}}(x_{t-1})-\tilde{g}_{\mathcal{I}_{t}}(\tilde{x})+\tilde{g}$. 
\item Set $x_{t}=\textrm{Pro}\textrm{x}_{\gamma_{s}R}$$\left(x_{t-1}-\gamma_{s}\,v_{t}\right)$. 
\end{enumerate}
\item Compute $\tilde{x}_{s}=\frac{1}{M_{s}}\sum_{t=1}^{M_{s}}x_{t}$.
\item Set $x_{0}=x_{M_{s}}$.
\end{enumerate}
\end{enumerate}
\end{algorithm}

\subsection{Convergence Rates}

We now state our main results on the convergence rate of Algorithm
1, and analysis on how it can effectively reduce the variance of gradient
estimation. The detailed proofs of all the techinical lemmas, theorems
and corollaries are referred to the supplymental material of this
main paper. 

From Algorithm 1, we know that the variance of the gradient estimation
comes from two sources, the randomness of selecting the component
function and estimation of smoothed function by sampling. The first
kind of variance comes from the stochastic selection of the gradient
of the smoothed component function, defined for a fixed $s$, as
\[
\epsilon_{t}^{1}:=\mathbb{E}\|\nabla\mathbb{E}\left[f_{i}(x_{t-1}+a_{s}\,Z)\right]-\nabla F_{a_{s}}(x_{t-1})\|_{2}^{2},
\]
and the second kind of variance comes from the estimation of the gradient
of smoothed component function by random sampling. Let $\mathcal{F}_{t}$ denote the $\sigma$-field
of the random variables $v_{t}$. The error is defined as
\[
\epsilon_{t}^{2}:=\mathbb{E}_{\mu}\|v_{t}-\nabla\mathbb{E}\left[f_{\mathcal{I}_{t}}(x_{t-1}+a_{s}\,Z)\right]|\mathcal{F}_{t-1}\|_{2}^{2},
\]
Define $e_{t}=v_{t}-\nabla\mathbb{E}\left[f_{\mathcal{I}_{t}}(x_{t-1}+a_{s}\,Z)\right],$
then $\epsilon_{t}^{2}:=\mathbb{E}_{\mu}\|e_{t}|\mathcal{F}_{t-1}\|_{2}^{2}$. By the
inequality of Arithmetic mean and Quadratic mean, we have
\begin{align*}
 & \mathbb{E}\left[\mathbb{E}_{\mu}\left[\|v_{t}-\nabla F_{a_{s}}(x_{t-1})|\mathcal{F}_{t-1}\|_{2}^{2}\right]\right]\\
\leq & 2\left\{ \mathbb{E}_{\mu}\|v_{t}-\nabla\mathbb{E}_{\mu}\left[f_{\mathcal{I}_{t}}(x_{t-1}+a_{s}\,Z)\right]|\mathcal{F}_{t-1}\|_{2}^{2}\right\} \\
 & +2\left\{ \mathbb{E}\|\nabla\mathbb{E}_{\mu}\left[f_{i}(x_{t-1}+a_{s}\,Z)\right]-\nabla F_{a_{s}}(x_{t-1})\|_{2}^{2}\right\} .
\end{align*}
We have following assumptions used throughout our convergence results. 
\begin{assumption}
\label{Assumption 2.1} (i) Set $\mathcal{X}:=\textrm{dom}(R)$. The
random variable $Z$ is zero-mean with density $\mu$ (with respect
to Lebesgue measure on the affine hull $\textrm{aff}(\mathcal{X})$
of $\mathcal{X}$). There are constants $L_{0}$ and $L_{i},\,i=1,...,N$
such that for $a>0$, $F_{a}(x):=\mathbb{E}_{\mu}\left[F(x+aZ\right]\leq F(x)+L_{0}a$,
and $\mathbb{E}\left[f_{i}(x+aZ)\right]$ has $\frac{L_{i}}{a}$-Lipschitz
continuous gradient with respect to the norm $\|\cdot\|.$ Additionally,
for $P$-a.e. $\mathcal{I}_{t}\in\left\{ 1,\ldots,\,N\right\} $,
the set $\textrm{dom}f_{i}(\cdot)\supseteq a_{0}\textrm{supp}\mu+\mathcal{X}$. 

(ii) The parameter $a_{s}$ that is exponentially decreasing w.r.t
$s$. i.e. $a_{s}=a_{0}\phi^{s}$ with $a_{0}>0$ and $0<\phi<1$. 
\end{assumption}
Assumption \ref{Assumption 2.1} (i) follows directly from the assumption in (\cite[Assumption A]{Duchi2012}).
From Assumption \ref{Assumption 2.1} (i), we know that $\mathbb{E}\left[F(x+aZ)\right]$
has $\frac{L_{1}}{a}$-Lipschitz continuous gradient with respect
to the norm $\|\cdot\|$, with $L_{1}\geq(1/N)\sum_{i=1}^{N}L_{i}$.The
function $F_{a}$ is guaranteed to be smooth whenever $a$ (and hence
$a_{s}$) is a density with respect to Lebesgue measure, so Assumption
\ref{Assumption 2.1} (i) ensures that $F_{a}$ is uniformly close
to $F$ and not too ``jagged''. Many smoothing distributions, including
Gaussians and uniform distributions on norm balls, satisfy Assumption
\ref{Assumption 2.1} (see Appendix in supplymental material). The containment of $a_{0}\textrm{supp}\mu+\mathcal{X}$
in $\textrm{dom}f_{i}(\cdot;\xi)$ guarantees that the subdifferential
$\partial f_{i}(\cdot)$ is nonempty at all sampled points $\tilde{x}+a_{s}Z$.
Indeed, since $\mu$ is a density with respect to Lebesgue measure
on $\textrm{aff}(\mathcal{X})$ , with probability 1 $\tilde{x}+a_{s}Z\in\textrm{relint}\textrm{dom}f_{i}(\cdot)$,
and thus the subdifferential $\partial f_{i}(\tilde{x}+a_{s}Z)\neq\emptyset$. 

Following Assumption impose condition on $\mu$ and error $e_{t}$
\begin{assumption}
\label{Assumption 2.2} (i) Given
any fixed $s$, there exists $B>0$ such that $\mathbb{E}_{\mu}\|e_{t}|\mathcal{F}_{t-1}\|_{2}^{2}\leq B,\,\forall t\in\{1,\,...,\,M_{s}\}$,
for any distribution $\mu$ following Assumption \ref{Assumption 2.1}. 

(ii) (\cite[Assumption B]{Duchi2012}) (sub-Gaussian errors) The error $e_{t}$ is $(\|\cdot\|_{\ast},\,\sigma)$
sub-Gaussian for some $\sigma>0$, meaning that, given fixed $s$,
with probability one
\[
\mathbb{E}\left[\exp(\|e_{t}\|_{\ast}^{2}/\sigma^{2})|\mathcal{F}_{t-1}\right]\leq\exp(1),\,\forall t\in\{1,\,...,\,M_{s}\}.
\]
\end{assumption}
\begin{lem}
\label{Lemma 2.2} Define the randomized smoothing (\ref{Smoothing}),
and suppose Assumption \ref{Assumption 2.1} holds, then we have $F(x)\leq F_{a_{s}}(x)\leq F_{a_{s-1}}(x),$
and $F(x)\geq F_{a_{s}}(x)-L_{0}\,a_{s}.$
\end{lem}
We have the following lemma and corollary give bounds on the variance
of the modified stochastic gradient $v_{t}$. To simplify, we use $\mathbb{E}_{\mu}[\cdot]$ to relpace $\mathbb{E}_{\mu}[\cdot|\mathcal{F}_{t-1}]$.
\begin{lem}
\label{Lemma 2.3} (Bounding the variance) Consider $P(x)$ as defined
in (\ref{Main problem}). Suppose Assumptions \ref{Assumption 1.1},
\ref{Assumption 2.1}, and \ref{Assumption 2.2} hold, and let $x_{\ast}=\arg\min_{x}P(x)$.
Then, in each iteration $s$, we have
\begin{align*}
 & \frac{1}{N}\sum_{i=1}^{N}\|\mathbb{E}_{\mu}\left[\tilde{g}_{i}(x)\right]-\mathbb{E}_{\mu}\left[\tilde{g}_{i}(x_{\ast})\right]\|_{2}^{2}\\
\leq & \frac{2L_{1}}{a_{s}}\left[P(x)-P(x_{\ast})\right]+2L_{0}L_{1}.
\end{align*}
\end{lem}
\begin{cor}
\label{Corallary 2.4} Consider $v_{t}$ in Algorithm 1. Conditioned
on $x_{t-1}$, and iteration $s$, we have $\mathbb{E}\left[\mathbb{E}_{\mu}[v_{t}]\right]=\nabla F_{a_{s}}(x_{t-1})$
, and
\begin{align}
 & \mathbb{E}\left[\mathbb{E}_{\mu}\left[\|v_{t}-\nabla F_{a_{s}}(x_{t-1})\|_{2}^{2}\right]\right]\nonumber \\
\leq & \frac{16L_{1}}{a_{s}}\left[P(x_{t-1})-P(x_{\ast})+P(\tilde{x})-P(x_{\ast})\right]\nonumber \\
 & +32L_{0}L_{1}+2\mathbb{E}_{\mu}\|e_{t}\|_{2}^{2}.\label{Variance}
\end{align}
\end{cor}
Based on the inequality (\ref{Variance}), when both $x_{t-1}$ and
$\tilde{x}$ converge to $x_{\ast},$the variance of $v_{t}$ is also
reduced to a constant value. As a result, the modified stochastic
gradient obtain much faster convergence rate than normal stochastic
gradient. 
\begin{thm}
\label{Theoerm 2.6} (Convergence rate) Suppose Assumptions \ref{Assumption 1.1},
\ref{Assumption 2.1}, and \ref{Assumption 2.2} hold, and let $x_{\ast}=\arg\min_{x}P(x)$,
and \textup{$x^{\phi}$ denote the initial solution.} Then under following
settings, (i) $\gamma_{s}=\frac{a_{s}}{25L}$; (ii) $\phi=1/8$; (iii)
$\mathbb{E}_{\mu}\|e_{t}|\mathcal{F}_{t-1}\|_{2}^{2}\leq B$, for
any $t$, the Algorithm 1 converges linearly $\mathbb{E}P[\tilde{x}_{s}]-P[x_{\ast}]\leq\left(\frac{1}{2}\right)^{s}D$,
where,
\begin{align}
D= & 2(P[x^{\phi}]-P[x_{\ast}])+\frac{25L_{1}\|x^{\phi}-x_{\ast}\|_{2}^{2}}{a_{0}\cdot M}\nonumber \\
 & +3L_{0}a_{0}+\frac{a_{0}\cdot M\cdot B}{24L_{1}}.\label{D}
\end{align}
In addition, Algorithm 1 has the time complexity $O(\log_{2}(D/\epsilon)$
, and a gradient complexity of $O(N\cdot m\cdot\log(D/\epsilon)+M\cdot D/\epsilon)$. 
\end{thm}
From Theorem \ref{Theoerm 2.6}, we know that Algorithm 1 converges
linearly with a robust rate $1/2$, which is independent of the parameters
in the algorithm as well as the Lipschitz modulus on smoothed objective
function and gradients. But the time complexity and gradient complexity
are both increased if these parameters and modulus as well as the
distance of the initial solution to the optimal solution are increased. 

Based on Theorem \ref{Theoerm 2.6}, we have the following novel high-probability
bound, where the randomness comes from both from incremental gradient
selection in SVRG and randomized smoothing. We give the detailed proof
procedure in Section 3. 
\begin{thm}
\label{Theorem 2.7} Suppose Assumptions \ref{Assumption 1.1}, \ref{Assumption 2.1},
and \ref{Assumption 2.2} hold. Then for any $\epsilon>0$ and $\delta_{1},\,\delta_{2}\in(0,\,1)$,
we have, $\mathbb{P}(P[\tilde{x}_{s}]-P[x^{\ast}]\leq\epsilon)\geq(1-\delta_{1})(1-\delta_{2})$,
provided that the number of stages $s$ satisfies, $s\geq\log\left(\frac{D^{\prime}}{\delta_{1}\,\epsilon}\right)/\log\left(2\right)$,
where $D^{\prime}$ is defined by the following equation,
\begin{align*}
D^{\prime}= & 2(P[x^{\phi}]-P[x_{\ast}])+\frac{25L_{1}\|x^{\phi}-x_{\ast}\|_{2}^{2}}{a_{0}\cdot M}\\
 & +3L_{0}a_{0}+\frac{a_{0}\cdot M\cdot B}{24L_{1}}\\
 & +\frac{a_{0}}{24L_{1}}\max\left\{ 8\sigma^{2}\log\frac{1}{\delta_{2}},\,12\sigma^{2}\sqrt{M\log\frac{1}{\delta_{2}}}\right\} .
\end{align*}
\end{thm}

\subsection{Remarks}

In this section, we bound the gradient estimation error $e_{t}$ by
properly choose the probability distribution $\mu$ for randomized
smoothing. We now turn to various corollaries of the above theorems
and the consequential optimality guarantees of the algorithm. More
precisely, we establish concrete convergence bounds for algorithms
using different choices of the smoothing distribution $\mu$. We begin
with a corollary that provides bounds when the smoothing distribution
$\mu$ is uniform on the $l_{2}$-ball. The conditions on $f_{i}$
in the corollary hold, for example, when $f_{i}(\cdot)$ is $L_{0}$-Lipschitz
with respect to the $l_{2}$-norm for $P$-a.e. sample of $i$. 
\begin{cor}
\label{Corollary 2.7} Suppose Assumptions \ref{Assumption 1.1} and
\ref{Assumption 2.1} hold. Let $\mu$ be uniform on $\mathcal{B}_{2}(0,\,1)$
and assume for each $s$, $\mathbb{E}\left[\|\partial f_{i}(x)\|_{2}^{2}\right]\leq L_{0}^{2}$
for $x\in\mathcal{X}+\mathcal{B}_{2}(0,\,a_{s})$, With stepsize $\gamma_{s}=\frac{a_{s}}{25L}$
and $\phi=1/8$, the algorithm converges linearly that $\mathbb{E}P[\tilde{x}_{s}]-P[x_{\ast}]\leq\left(\frac{1}{2}\right)^{s}D$,
where,
\begin{align}
D= & 2(P[x^{\phi}]-P[x_{\ast}])+\frac{25L_{0}\sqrt{d}\|x^{\phi}-x_{\ast}\|_{2}^{2}}{a_{0}\cdot M}\nonumber \\
 & +3L_{0}a_{0}+\frac{a_{0}\cdot M\cdot L_{0}}{24m\sqrt{d}}.\label{D-1}
\end{align}
\end{cor}
The following corollary shows convergence rate when smoothing with
the normal distribution.
\begin{cor}
\label{Corollary 2.8} Suppose Assumptions \ref{Assumption 1.1} and
\ref{Assumption 2.1} hold. Let $\mu$ be the $d$-dimensional normal
distribution with zero mean and identity covariance $I$ and assume
that $f_{i}(\cdot)$ is $L_{0}$-Lipschitz with respect to the $l_{2}$-norm
for $P$-a.e. $i$. With stepsize $\gamma_{s}=\frac{a_{s}}{25L}$
and $\phi=1/8$, the algorithm converges linearly that $\mathbb{E}P[\tilde{x}_{s}]-P[x_{\ast}]\leq\left(\frac{1}{2}\right)^{s}D$,
where,
\begin{align}
D= & 2(P[x^{\phi}]-P[x_{\ast}])+\frac{25L_{0}\|x^{\phi}-x_{\ast}\|_{2}^{2}}{a_{0}\cdot M}\nonumber \\
 & +3L_{0}\sqrt{d}a_{0}+\frac{a_{0}\cdot M\cdot L_{0}}{24m}.\label{D-1-1}
\end{align}
\end{cor}
The following corollary shows convergence rate when smoothing with
the uniform distribution on the $l_{\infty}$-ball.
\begin{cor}
\label{Corollary 2.9} Suppose Assumptions \ref{Assumption 1.1} and
\ref{Assumption 2.1} hold. Let $\mu$ be uniform on $\mathcal{B}_{\infty}(0,\,1)$
and and assume that $f_{i}(\cdot)$ is $L_{0}$-Lipschitz with respect
to the $l_{1}$-norm over\textup{ $\mathcal{X}+\mathcal{B}_{2}(0,\,a_{s})$
for} $P$-a.e. $i$. With stepsize $\gamma_{s}=\frac{a_{s}}{25L}$
and $\phi=1/8$, the algorithm converges linearly that $\mathbb{E}P[\tilde{x}_{s}]-P[x_{\ast}]\leq\left(\frac{1}{2}\right)^{s}D$,
where,
\begin{align}
D= & 2(P[x^{\phi}]-P[x_{\ast}])+\frac{25L_{0}\|x^{\phi}-x_{\ast}\|_{2}^{2}}{a_{0}\cdot M}\nonumber \\
 & +\frac{3d}{2}L_{0}a_{0}+\frac{a_{0}\cdot M\cdot L_{0}}{6m}.\label{D-1-2}
\end{align}
\end{cor}
Corollaries \ref{Corollary 2.7}, \ref{Corollary 2.8} and \ref{Corollary 2.9}
show the robust convergence rate $1/2$ as well as the complexity
results based on problem dimension $d$ and number of samples $m$.
From (\ref{D-1-1}), it shows that the complexity of Algorithm 1 is
$O(\log_{2}((m\sqrt{d}+1)/m\epsilon))$ which is increasing with $\sqrt{d}$
and decreasing with $m$. From (\ref{D-1-2}), it shows that the complexity
of Algorithm 1 is $O(\log_{2}((md+1)/m\epsilon))$ which is also increasing
with $d$ and decreasing with $m$. Compared with the case when $\mu$
following the normal distribution with zero mean and identity covariance
$I$, the complexity grows fast with respect to the growth of dimension
$d$; while, from (\ref{D-1}), it shows that that the complexity
of Algorithm 1 is $O(\log_{2}((dm+1)/\sqrt{d}m\epsilon))$, which
is also decreasing with $m$ and decreasing with $d$, explicitly,
when $d\leq a_{0}^{2}M^{2}/(600m\|x^{\phi}-x_{\ast}\|_{2}^{2})$,
but increasing with $d$ when $d\geq a_{0}^{2}M^{2}/(600m\|x^{\phi}-x_{\ast}\|_{2}^{2})$.
The complexity results in Corollaries\ref{Corollary 2.7}, \ref{Corollary 2.8}
and \ref{Corollary 2.9} all reveal that increasing the size of sampling
perturbation random variable $Z$ can reduce the time complexity and
gradient complexity of the algorithm on high-dimensional problems. 

\section{Applications and Numerical Experiments}

In this section, we describe applications of our results and give
experiments that illustrate our theoretical predictions. 

\subsection{Two applications}

The first application is ranking problems in machine
learning. Described in (\cite{Agarwal2009}), one is given a finite number of examples of order relationships among
instances in some instance space $\mathcal{S}$, and the goal is to
learn from these examples a ranking or ordering over $\mathcal{S}$
that ranks accurately future instances. In the Bipartite ranking problem,
instances come from two categories, positive and negative; the learner
is given examples of instances labeled as positive or negative. Specifically
for our case, the learner is given a training sample $S=\{(x_{i},\,y_{i})\}_{i=\{1,..,N\}}$,
and we always let $x_{i}$ record the positive instance and $y_{i}$
the negative one. The goal is to learn a real-valued ranking function
$\rho:\,\mathcal{S}\rightarrow\mathbb{R}$, in which positive instances
are ranked higher than negative ones. The learning quality of a choice
function $\rho$ can be measured by certain loss function. In Bipartite
loss, we let $\rho(x_{i})\geq\rho(y_{i})$ give the loss to be zero
and $\rho(x_{i})<\rho(y_{i})$ denote the violation level for the
mis-ordering for the prospect pair $(x_{i},\,y_{i})$ and the loss
is recorded as $1$. Then the function $\rho$ can be estimated by
solving following empirical risk minimization problem
\[
\min_{\rho}\,\frac{1}{N}\sum_{i=1}^{N}I_{\{\rho(x_{i})-\rho(y_{i})<0\}}+\lambda R(\rho),
\]
where $I_{\{t\}}$ is $1$ if $t$ is true and $0$ otherwise, $\lambda>0$
and $R$ is a convex regularization functional. In (\cite{Chen2012}),
the authors mentioned that due to the non-convexity of $I$, the empirical
minimization problem based on $I$ is NP-hard. Thus we may consider
replacing $I$ by a convex upper loss function e.g., Hinge loss. Then
the problem becomes
\[
\min_{\rho}\,\frac{1}{N}\sum_{i=1}^{N}\max\left\{ 1-(\rho(x_{i})-\rho(y_{i})),\,0\right\} +\lambda R(\rho),
\]
and popular choices of the regularization term $R$ include the $l_{1}$-norm
(Lasso), $l_{2}$-norm (ridge regression), and the linearly combination
of $l_{1}$ and $l_{2}$ norm (elastic net). Normally, if we consider
$\rho$ containing in Reproducing kernel Hilbert space (RKHS) with
kernel function $K$, and set $R(\rho)=\|\rho\|_{K}^{2}$. The problem
\[
\min_{\rho\in\mathcal{F}_{K}}\,\frac{1}{N}\sum_{i=1}^{N}\max\left\{ 1-(\rho(x_{i})-\rho(y_{i})),\,0\right\} +\lambda\|\rho\|_{K}^{2},
\]
is called Bipartite RankSVM Algorithm. In terms of solution methods, in (\cite{Herbrich2000,Joachims2002}),
the authors solve the problem by introducing slack variables and taking
the Lagrangian dual results in a convex quadratic program (QP), and
then use a standard QP solver to tackle the problem. However, when
$N$ is extremely large, the number of decision variables and constraints
of this quadratic program will also become extremely large, which
will cause computational efficiency. To improve, we could develop
a functional extension to our RS-SVRG algorithm, where we can refer
to the functional gradient proposed in (\cite{Kivinen2004}). Our algorithm
can achieve linear convergence with lower complexity than other gradient
based algorithms in large-scale machine learning problems. 

The second application is related to Similarity learning (or Metric
learning) problems seen in (\cite{Xing2003,Shai-Shwartz2004}), in which
the learner is given a set of $N$ points $\{a_{1},\,...,\,a_{N}\}\in\mathbb{R}^{d}$
and a matrix $B\in\mathbb{R}^{N\times N}$ indicating which points
are close together in an unknown metric. The goal is to estimate a
positive semi-definite matrix $X\succeq0$ such that $\left\langle (a_{i}-a_{j}),\,X(a_{i}-a_{j})\right\rangle $
is small when $a_{i}$ and $a_{j}$ belong to the same class or close,
while $\left\langle (a_{i}-a_{j}),\,X(a_{i}-a_{j})\right\rangle $
is large when $a_{i}$ and $a_{j}$ belong to different classes. It
is desirable that the matrix $X$ has low rank, which allows the statistician
to discover structure or guarantee performance on unseen data. As
a concrete illustration, suppose that we are given a matrix $C\in\{-1,\,1\}^{N\times N}$,
where $b_{ij}=1$ if $a_{i}$ and $a_{j}$ belong to the same class
and $c_{ij}=-1$ otherwise. In this case, one possible optimization-based
estimator involves solving the nonsmooth program developed as (\cite{Duchi2012}),
\begin{align*}
\min_{X,\,x}\quad & \frac{1}{\left(\frac{N}{2}\right)}\sum_{i<j}\left[1+c_{ij}(\textrm{tr}(X(a_{i}-a_{j})(a_{i}-a_{j})^{\top})+x\right]_{+}\\
 & +\lambda_{1}(\|X\|_{2}^{2}+\|x\|_{2}^{2})+\lambda_{2}(\|X\|_{1}+\|x\|_{1})\\
\textrm{s.t.}\quad & X\succeq0,\\
\quad & \textrm{tr}(X)\leq C,
\end{align*}
where the regularization term $\lambda_{1}(\|X\|_{2}^{2}+\|x\|_{2}^{2})+\lambda_{2}(\|X\|_{1}+\|x\|_{1})$
with non-negative $\lambda_{1}$ and $\lambda_{2}$ is called elastic
net, which overcomes the limitations of Lasso on ``large $d$, small
$N$'' case and on highly correlated data. The stochastic oracle
for this problem is simple: given a query matrix $X$, the oracle
chooses a pair $(i,\,j)$ uniformly at random and then returns the
subgradient of each component function as $\textrm{sign}\left[\left\langle (a_{i}-a_{j}),\,X(a_{i}-a_{j})\right\rangle -c_{ij}\right](a_{i}-a_{j})(a_{i}-a_{j})^{\top}$.
Obviously, we are able to apply our RS-SVRG to tackle this Similarity
learning problem. Here the positive semi-definite and sparsity constraints
can be transformed into projection term in the algorithm. 

\subsection{Experiment results}

In this section we present results of several numerical experiments
to illustrate the properties of the RS-SVRG method and compare its
performance with several related algorithms, on a simple ranking problem
in machine learning. We focus on the elastic net regularized ranking
problem with hinge loss and linear ranking function: given a set of
training example pairs $S=\{(x_{i},\,y_{i})\}_{i=\{1,..,N\}}$ with
orders, where $x_{i},y_{i}\in\mathbb{R}^{d}$. Here each element in
vectors $x_{i}$ and $y_{i}$ is randomly generated with support bounded
in $[0,\,100]$. In the simplest case, we assume that $\rho:\,\mathcal{S}\rightarrow\mathbb{R}$
are restricted to linear ranking functions i.e., $\rho(x)=w^{\top}x+b,\,w\in\mathbb{R}^{d},\,b\in\mathbb{R}$. 

We find the optimal predictor $w\in\mathbb{R}^{d}$ by solving
\begin{align*}
\min_{w\in\mathbb{E}^{d}}\, & \left\{ \frac{1}{N}\sum_{i=1}^{N}\max\left\{ 1-(x_{i}^{\top}w-y_{i}^{\top}w),\,0\right\} \right.\\
 & \left.+\lambda_{1}\|w\|_{2}^{2}+\lambda_{2}\|w\|_{1}\right\} .
\end{align*}
In our experiments, we generate $N=1000$ training example pairs,
the number of inner iterations $M=2$, and the total number of outer
iteration corresponding to $s$ equals $10$. 

\subsubsection{Comparison with related algorithms}

We implement the following algorithms to compare with our RS-SVRG. 
\begin{itemize}
\item Prox-SGD (Stochastic subgradient descent): The proximal stochastic
gradient method is given as: $x_{t}=\textrm{prox}_{R}\left\{ x_{t-1}-\gamma_{t}\,\partial f_{\mathcal{I}_{t}}(x_{t-1})\right\} $,
and $\mathcal{I}_{t}$ is drawn randomly from $\{1,\,...,\,N\}$.
Here we set the diminishing stepsize as $\gamma_{t}=1/\sqrt{t}$.
\item Prox-FGD (Full subgradient descent): The proximal full gradient method
is given as: $x_{t}=\textrm{prox}_{R}\left\{ x_{t-1}-\gamma\,\frac{1}{N}\sum_{i=1}^{N}\partial f_{i}(x_{t-1})\right\} $,
with a constant stepsize. 
\item RS-SGD (Randomized smoothing stochastic gradient descent): The algorithm
applies the same randomized smoothing technique in (\cite{Duchi2012}),
but instead of using the dual averaging scheme in (\cite{Xiao2010,Nesterov2005}),
we use normal proximal stochastic gradient descent with diminishing
stepsize for updating the solution. 
\item RS-SAG (Randomized smoothing stochastic average gradient descent):
Since the theory of SAG on nonsmooth objective has not been investigated,
RS-SAG applies the same randomized smoothing technique in (\cite{Duchi2012})
for smoothing the nonsmooth objective, and then use proximal stochastic
average gradient proposed in (\cite{Schmidt2017}) for updating the
solution. 
\item Prox-SDCA: See (\cite{Shalev-Shwartz2012,Shalev-Shwartz2013}), which obtains the
overall gradient complexity $O((n+1)/\epsilon))$ for nonsmooth objective. 
\item Accelerated RS-SGD: The algorithm applies randomized smoothing
technique in (\cite{Duchi2012}), and applies smooth minimization of
nonsmooth functions (Nesterov's smoothing) in (\cite{Nesterov2005})
for updating the solution. In (\cite{Nesterov2005}), Nesterov's smoothing
method improves the number of iterations of gradient schemes from
$O(1/\epsilon^{2})$ to $O(1/\epsilon)$. 
\item Bundle: Bundle method for large-scale machine learning problems in
(\cite{Le2008}), with complexity $O(1/\epsilon)$ for general convex
problems. 
\end{itemize}
We test our algorithm on three problems: Lasso ($\lambda_{1}=0,\,\lambda_{2}=0.01$),
ridge regularized ($\lambda_{1}=0.01,\,\lambda_{2}=0$) and elastic
net regularized ($\lambda_{1}=\lambda_{2}=0.01$) problem. We plot
the optimality gap $P(\tilde{x}_{s})-\min_{x}P(x)$ as a function
a number of iteration $s$. The convergence performance of our algorithm
compared with other gradient-based methods is shown in Figure 1, 2
and 3 . In these experiments, we choose $a_{0}=1$, $M=2$, the sampling
size $m=5$. The problem dimension $d=10$. We choose the distribution
of randomized smoothing to be $N(0,\,I_{d\times d})$ distribution.
Figure 1,2 and 3 show that our RS-SVRG converge to the optimum within
ten iterations on large-scale Lasso regularized, ridge regularized
and elastic net regularized problems, which is much superior than
Prox-SGD, Prox-FGD and RS-SGD methods. In addition, RS-SGD improves
the convergence rate compared with SGD, which shows the benefit of
randomized smoothing that transforms nonsmooth objective to smooth
one. Moreover, Figure 1,2 and 3 show that RS-SVRG and Prox-SDCA have
close convergence performance which is superior to Prox-SAG.

Table 1 and Figure 4 show the advantages of our RS-SVRG than other
state-of-art nonsmooth algorithms. Figure 4 illustrates that RS-SVRG,
Accelerated RS-SGD and Bundle method all converge extremely fast to
the optimum for a relatively small-scale problem ($d=10$, $S=10$,
$N=1000$), and Accelerated RS-SGD and Bundle method even performs
silently better than our method. Table 1 shows the advantages of our
method in terms of computational efficiency than the other methods.
We record the computational time of our ranking problem for several
value of $d$, $N$ and $S$. As the size of training data set $N$
and number of iterations $S$ grow, the Accelerated RS-SGD will have
relatively higher computational time compared to our method. Then
as the size of training data set $N$ and number of iterations $S$
grow, as well as the problem dimension $d$ grow, the computational
complexity of Bundle method will grow explosively. The reason is that
for Accelerated RS-SGD and Bundle method, it is required to store
all the historical subgradient information on each component function,
and then use this information to solve optimization problems with
dimension $d$. Instead, Our RS-SVRG don not need to record all the
historical gradient estimations, and has relatively stable computational
time even for solving problems with extremely large size of training
data, and high dimensions. 

\begin{figure}
\begin{centering}
\includegraphics[scale=0.2]{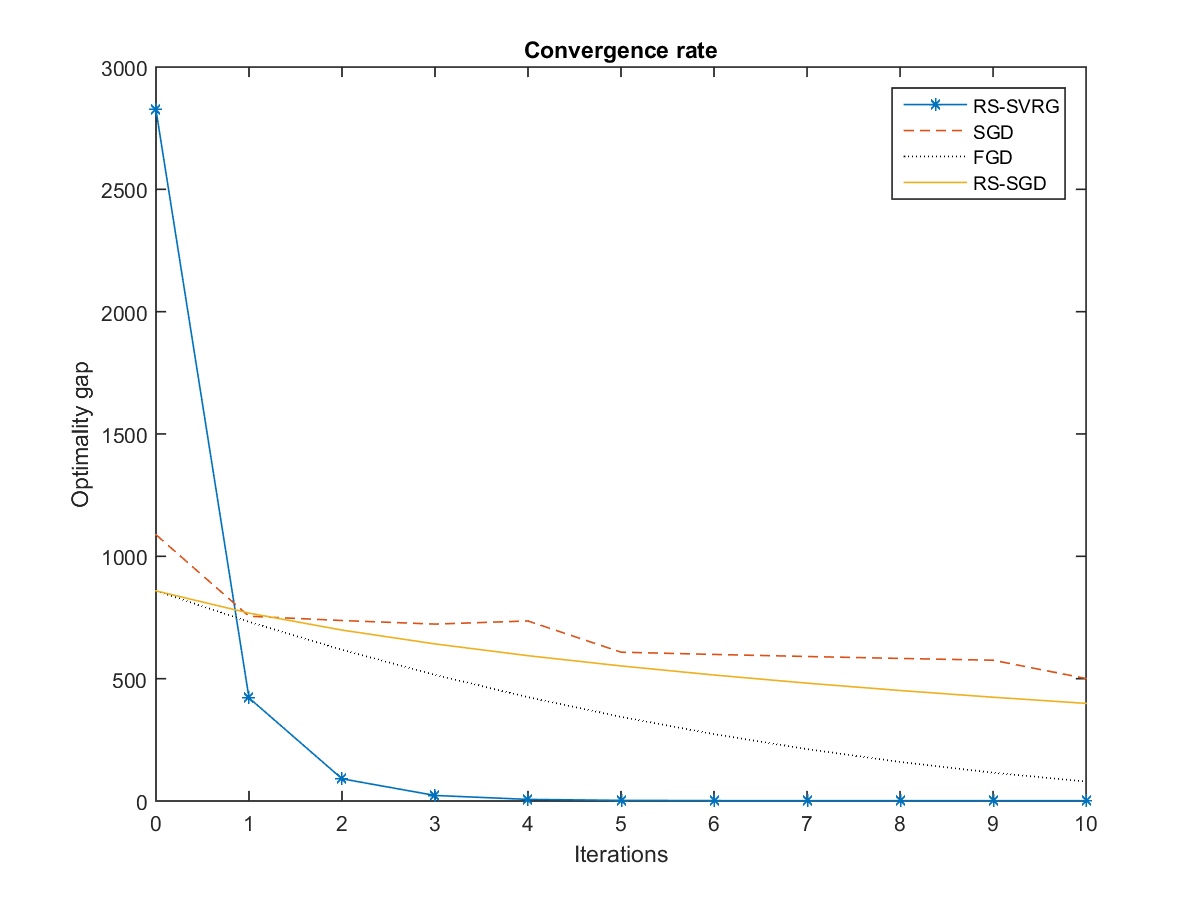}\includegraphics[scale=0.2]{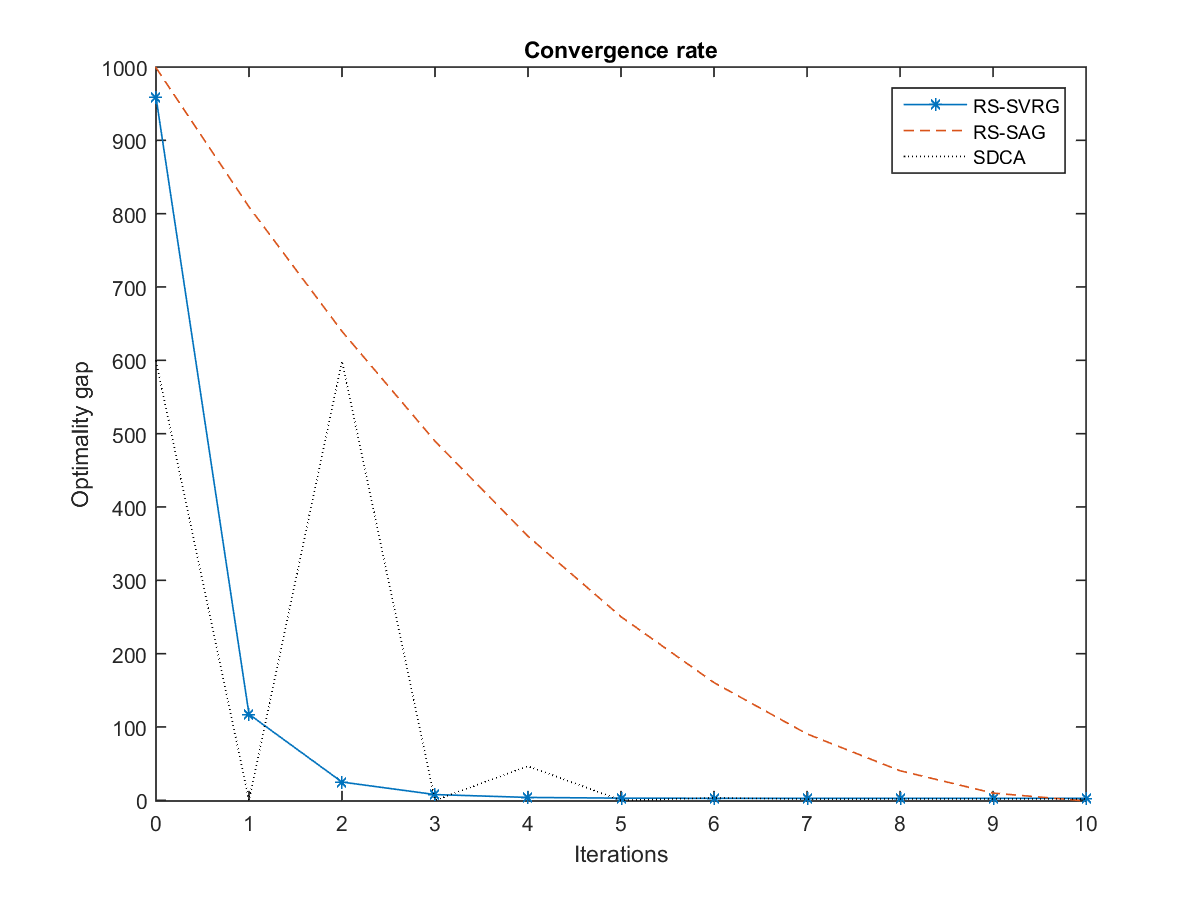}
\par\end{centering}
\caption{Ridge }
\end{figure}

\begin{figure}
\begin{centering}
\includegraphics[scale=0.2]{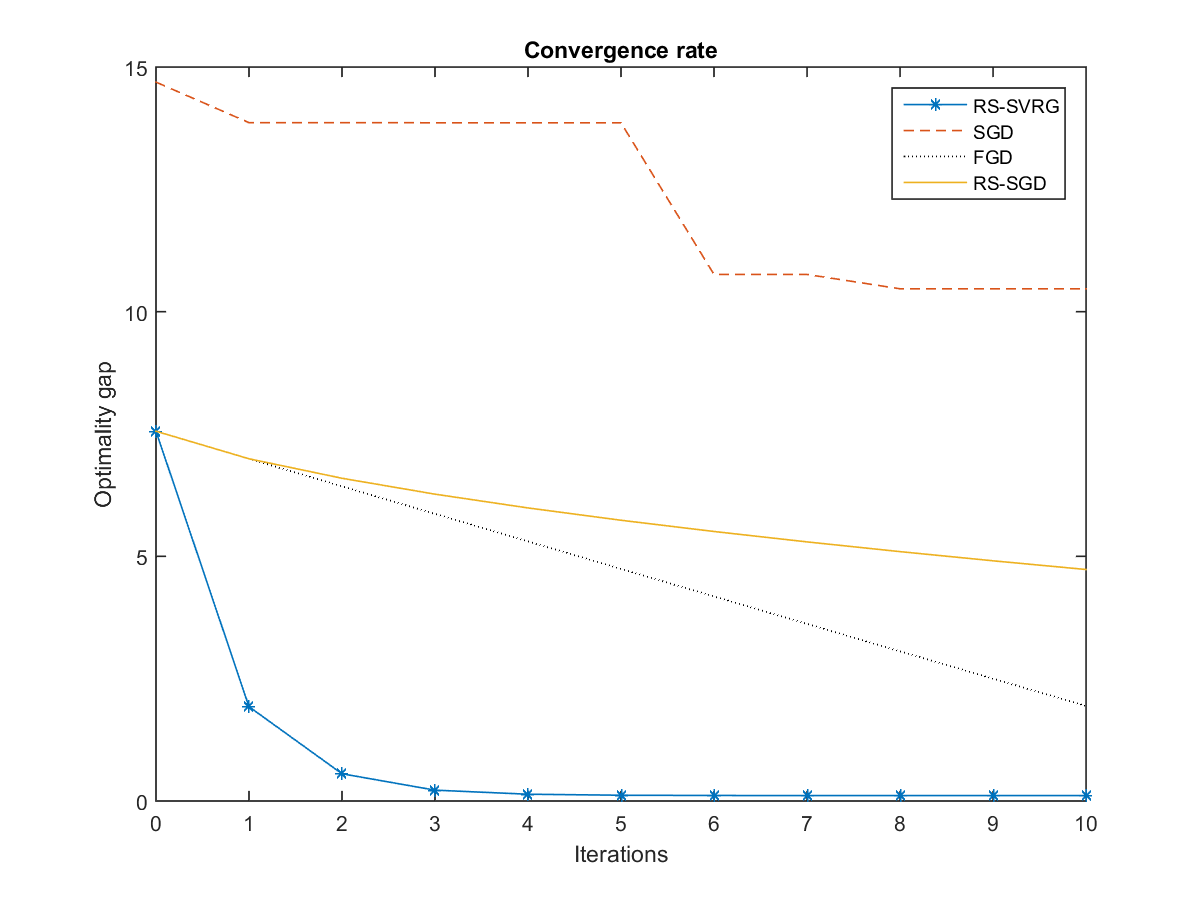}\includegraphics[scale=0.2]{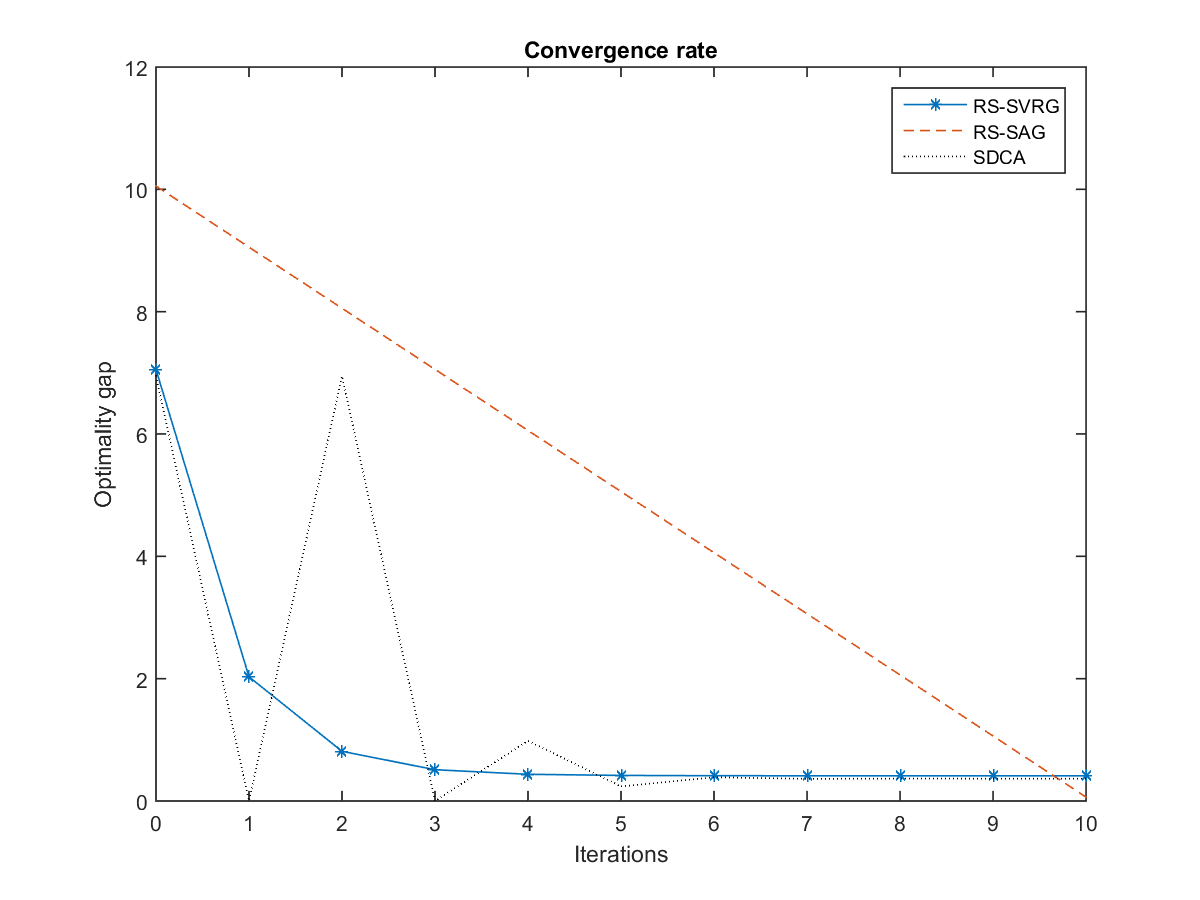}
\par\end{centering}
\caption{Lasso }
\end{figure}

\begin{figure}
\begin{centering}
\includegraphics[scale=0.2]{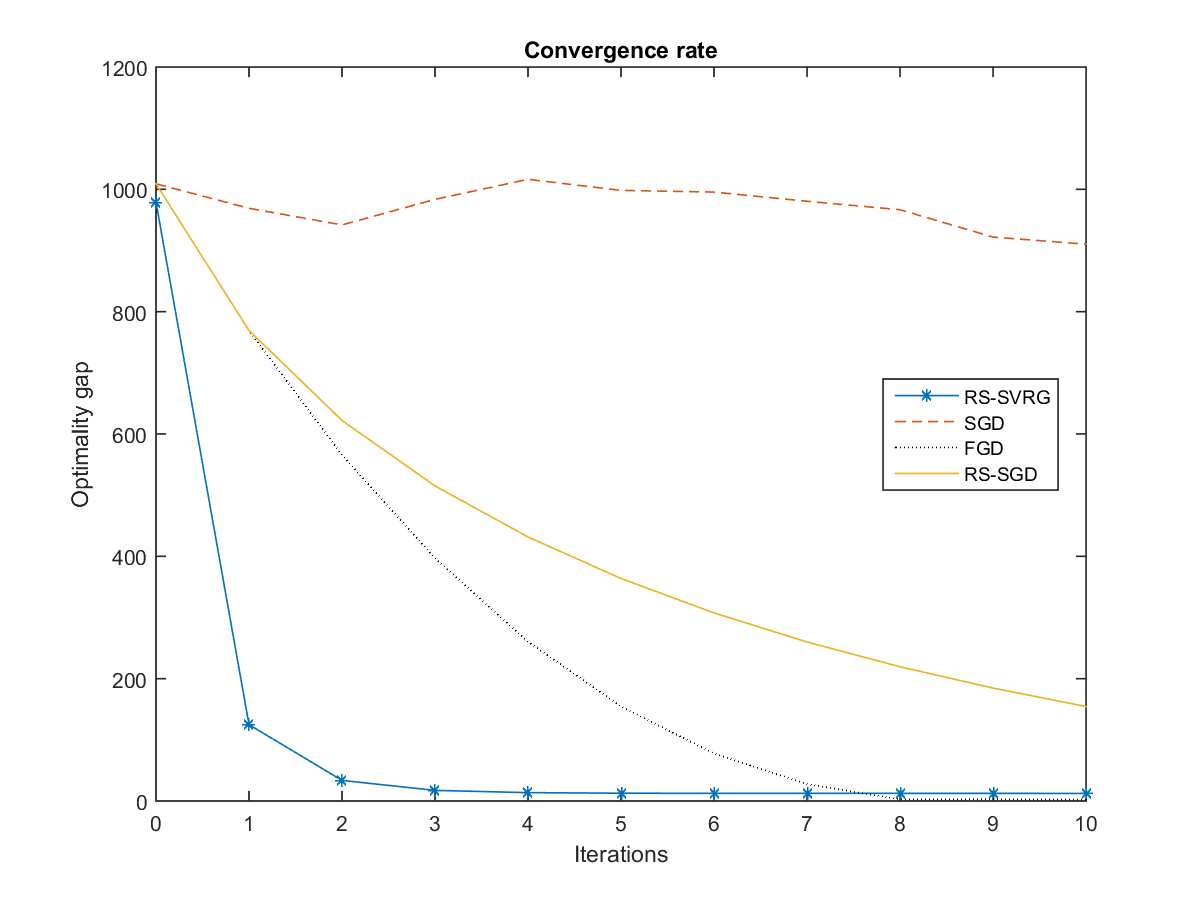}\includegraphics[scale=0.2]{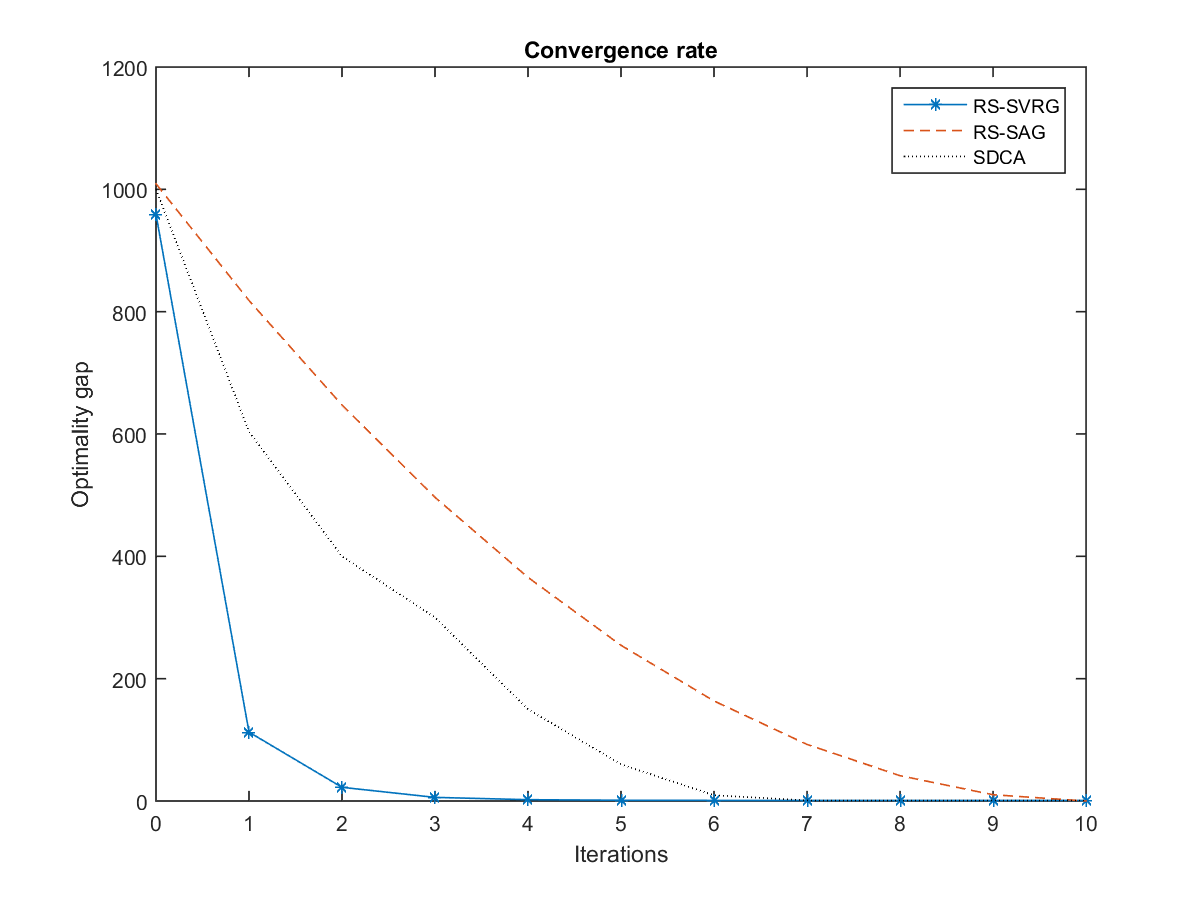}
\par\end{centering}
\caption{Elastic Net }
\end{figure}

\begin{figure}
\begin{centering}
\includegraphics[scale=0.4]{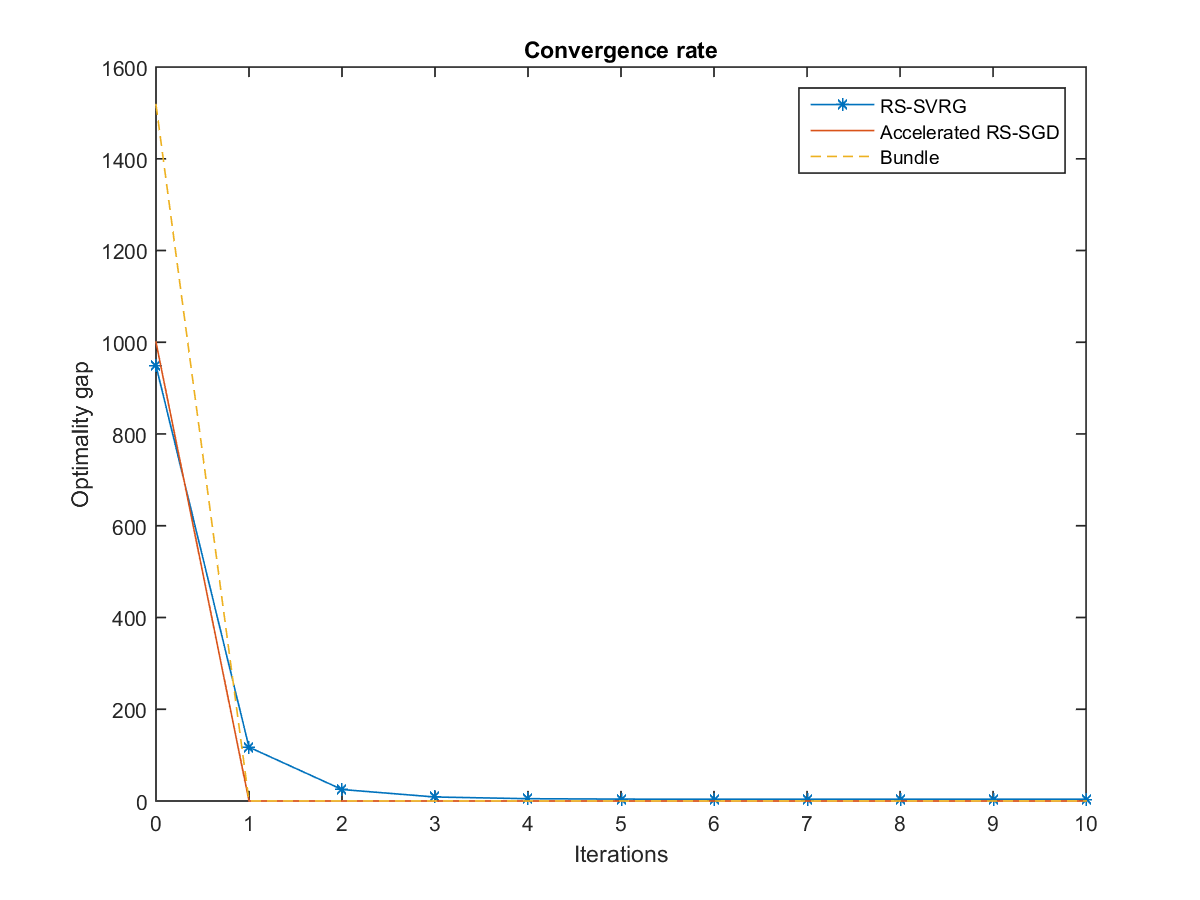}
\par\end{centering}
\caption{Comparison}
\end{figure}

\begin{table}
\begin{center}
	{\small{}}%
	\begin{tabular}{cccccc}
		\hline 
		{\small{}$d$} & {\small{}$N$} & {\small{}$S$} & {\small{}RS-SVRG} & {\small{}A-RS-SGD} & {\small{}Bundle}\tabularnewline
		\hline 
		{\small{}$10$} & {\small{}$1000$} & {\small{}$10$} & {\small{}7.379s} & {\small{}1.102s} & {\small{}5.734s}\tabularnewline
		{\small{}$200$} & {\small{}$1000$} & {\small{}$10$} & {\small{}8.815s} & {\small{}1.217s} & {\small{}12.214s}\tabularnewline
		{\small{}$10$} & {\small{}$5000$} & {\small{}$100$} & {\small{}15.795s} & {\small{}23.242s} & {\small{}54.165s}\tabularnewline
		{\small{}$200$} & {\small{}$5000$} & {\small{}$100$} & {\small{}18.702s} & {\small{}23.022s} & {\small{}268.438s}\tabularnewline
		{\small{}$10$} & {\small{}$10000$} & {\small{}$10$} & {\small{}2.453s} & {\small{}2.767s} & {\small{}60.965s}\tabularnewline
		{\small{}$500$} & {\small{}$10000$} & {\small{}$10$} & {\small{}5.131s} & {\small{}5.349s} & {\small{}6562.426s}\tabularnewline
		\hline 
	\end{tabular}
	\par\end{center}{\small \par}
\caption{Computational Time}
\end{table}

\subsubsection{Effects of sampling and problem dimension}

In this section, we focus on ridge regularized problem, i.e., $\lambda_{1}=0.01,\,\lambda_{2}=0$
with different randomized sampling size $m$ and different problem
dimension $d$. We study the effects of sampling and problem dimension
under different choice of random sampling distribution $\mu$. Figure
5, show the results as predicted by our theory and discussion in Section
2, receiving more samples $m$ gives improvements in reducing the
optimality gap as a function of iteration. The plots $m=50$ and $m=100$
are essentially indistinguishable, which implies that when the sampling
size is large enough, there should be no obvious improvement in actual
iteration taken to minimize the objective. Our theory also predicts
that the upper bound for the convergence rate contains a term with
$O(1/m$) from Corollaries \ref{Corollary 2.7}, \ref{Corollary 2.8}
and \ref{Corollary 2.9}, which means that the improvement of optimality
gap is deceasing as $1/m$ approaching zero. Figure 6 and 7 illustrate
the performance of the algorithm with respect to the problem dimension
$d$ when $\mu$ follows $N(0,\,I_{d\times d})$, and follows uniform
distribution on $\mathcal{B}_{\infty}(0,\,1)$, respectively. The results reveal
that the optimality gap increases significantly with the increase
of problem dimension. Besides, Figure 6 shows the optimality gap grows
approximately in linear relationship with $\sqrt{d}$, conforming
to what we have stated in Corollaries \ref{Corollary 2.8}. Figure
7 shows the optimality gap grows strictly linear with $d$, conforming
to what we have stated in Corollaries \ref{Corollary 2.9}.

\begin{figure}
\begin{centering}
\includegraphics[scale=0.4]{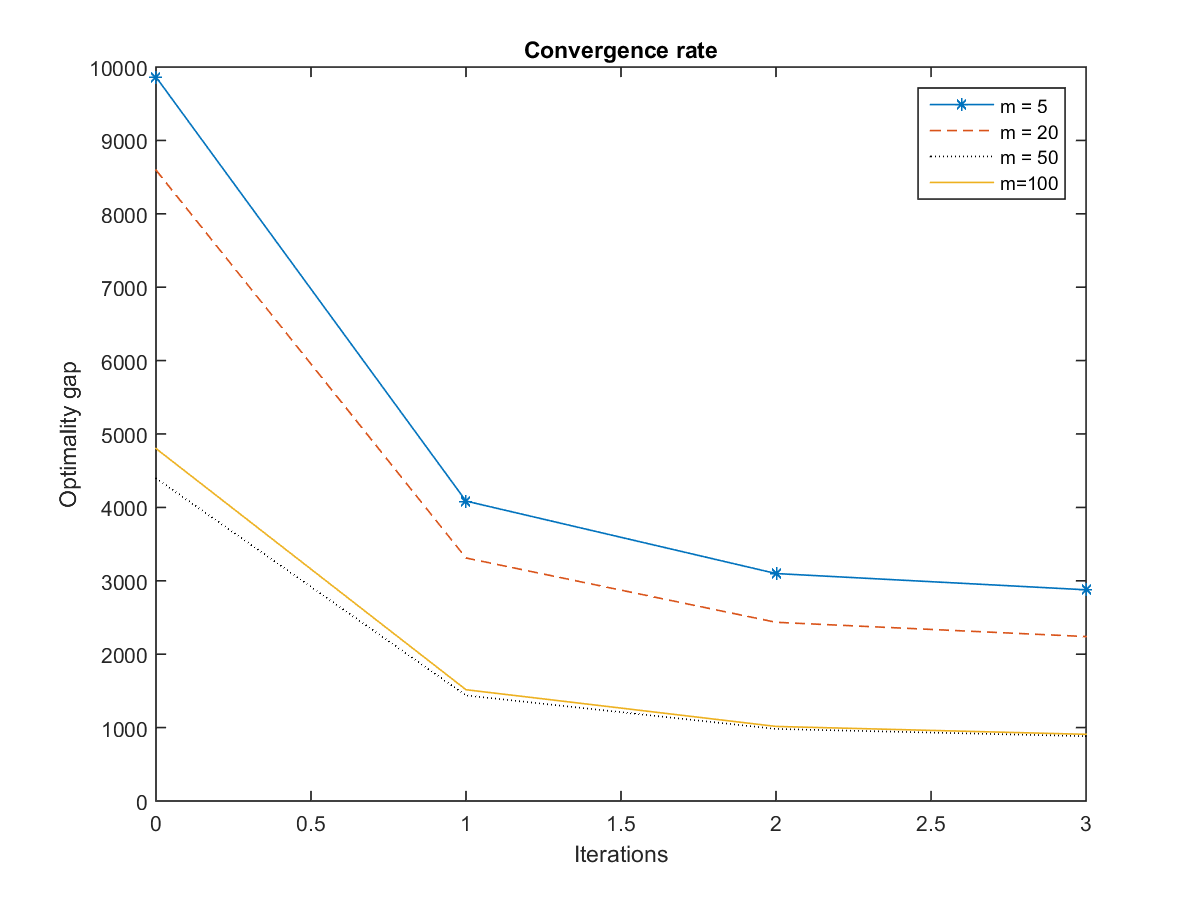}
\par\end{centering}
\caption{Effect of Samplings}
\end{figure}

\begin{figure}
\begin{centering}
\includegraphics[scale=0.2]{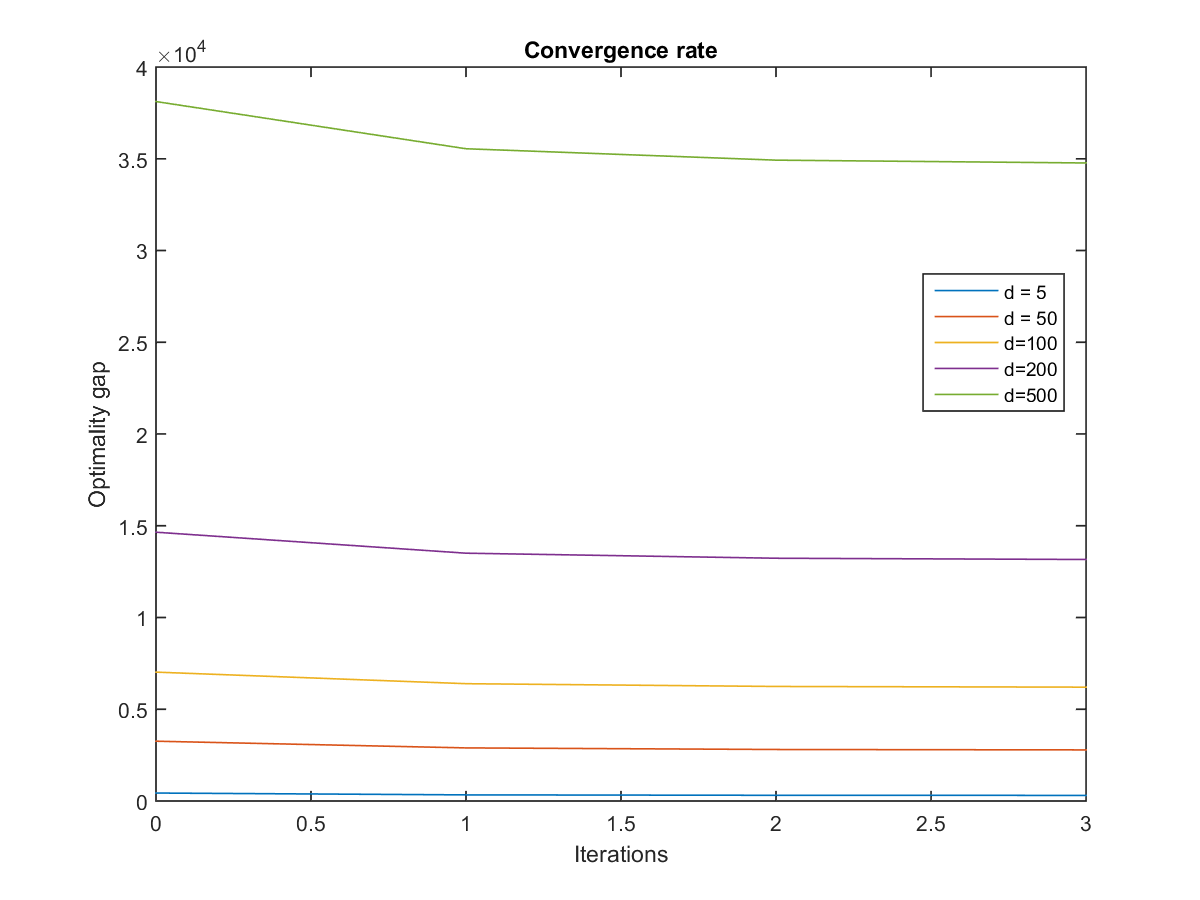}\includegraphics[scale=0.2]{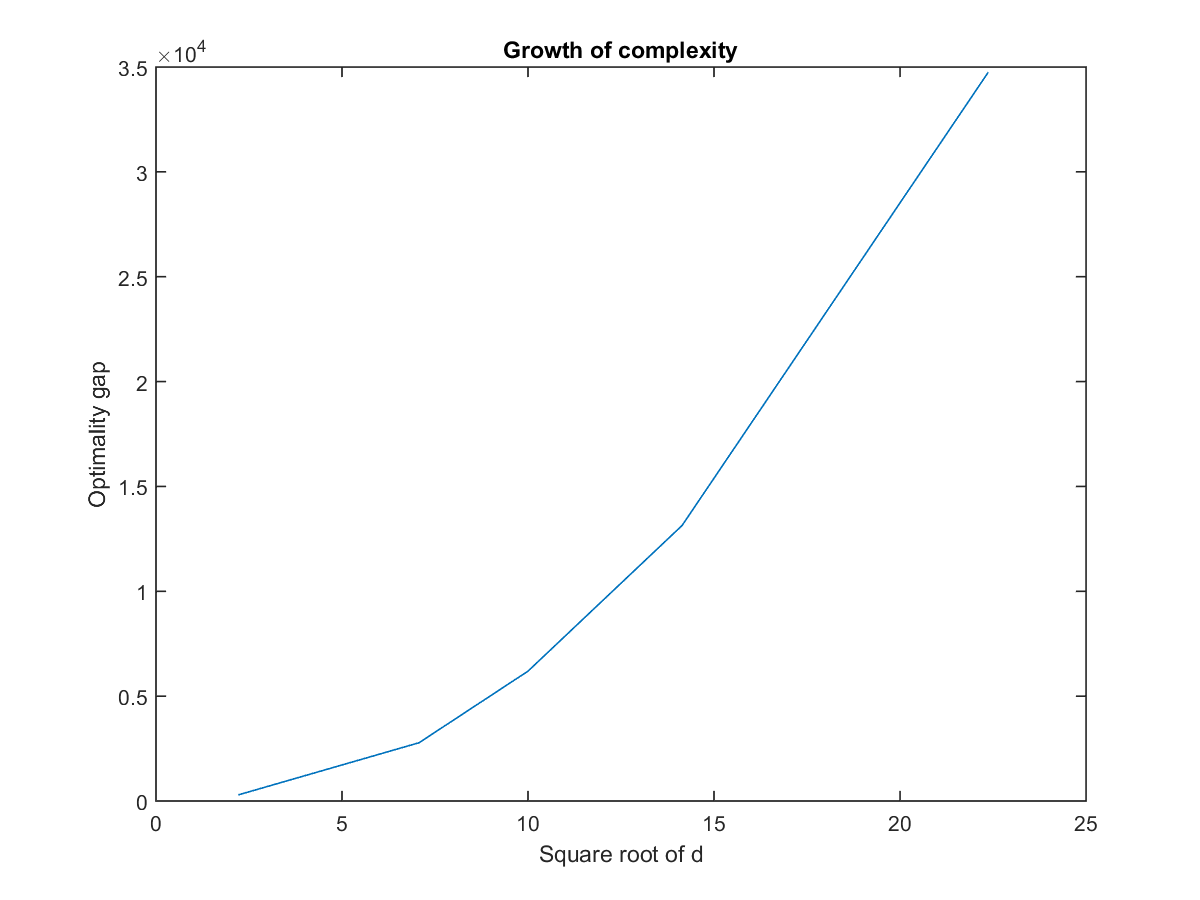}
\par\end{centering}
\caption{Effect of Dimension-$N(0,\,I_{d\times d})$}
\end{figure}

\begin{figure}
\begin{centering}
\includegraphics[scale=0.2]{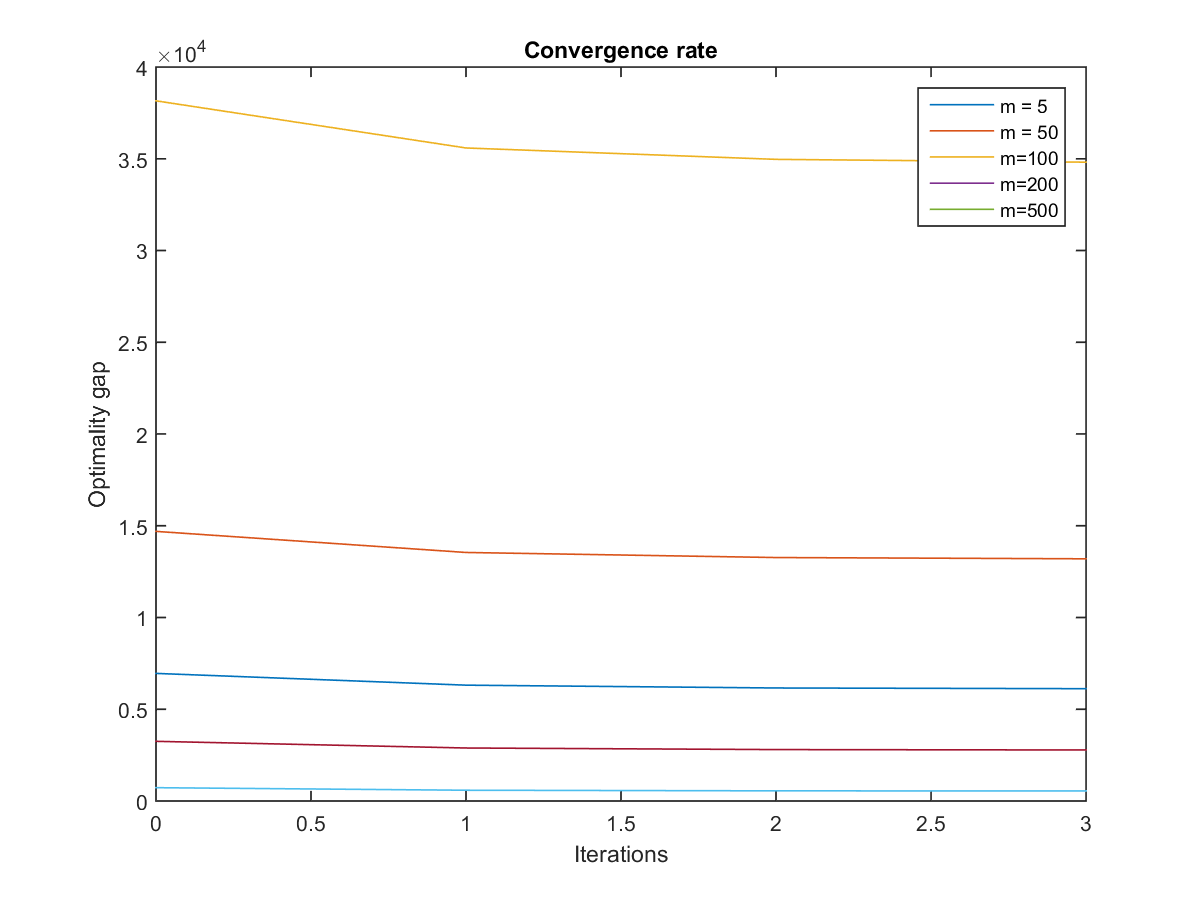}\includegraphics[scale=0.2]{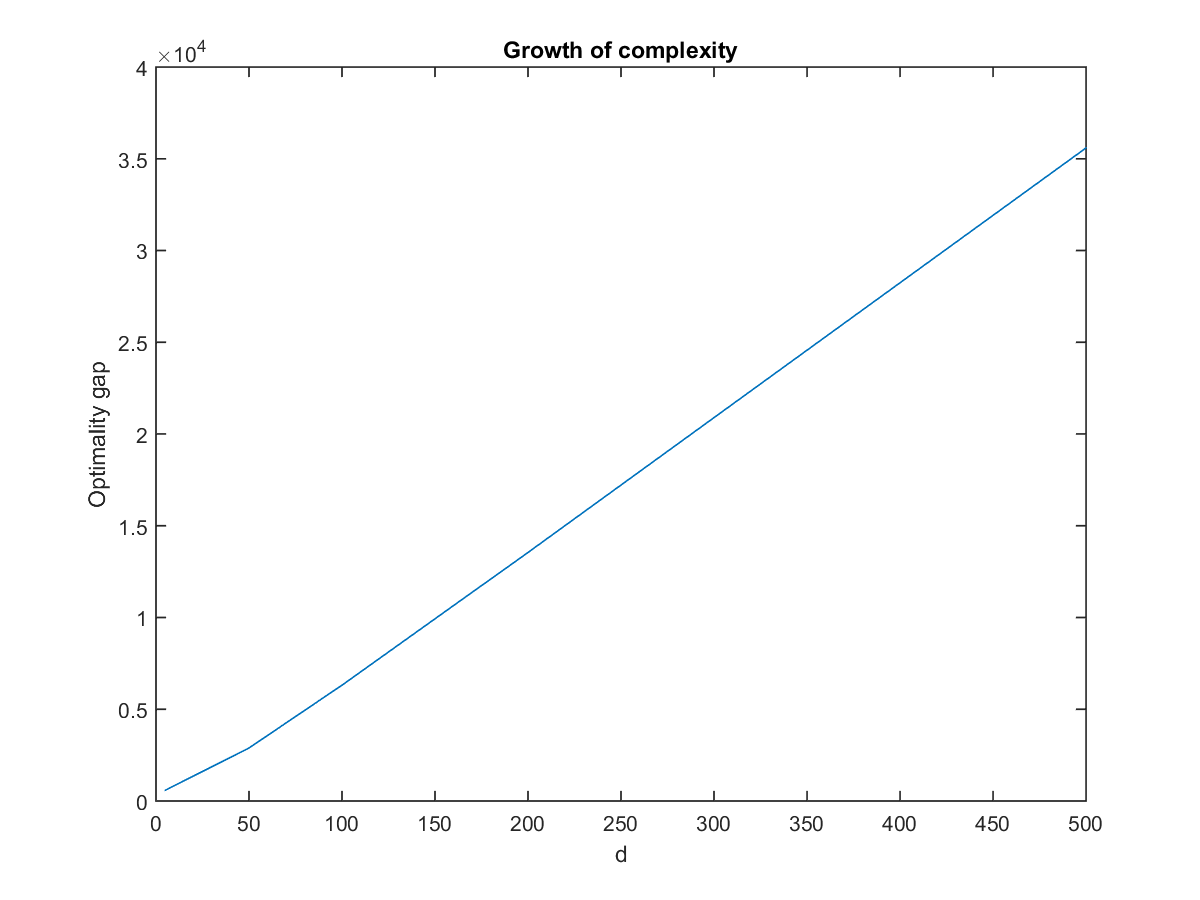}
\par\end{centering}
\caption{Effect of Dimension-Uniform on $B_{\infty}(0,\,1)$}
\end{figure}

\section{Conclusion}

In this paper, we develop and analyze a new method, for minimizing the sum of two convex functions: one is the average of a large number of convex component functions (not necessarily smooth and strongly-convex), and the other is a general convex function that admits a simple proximal mapping. Our RS-SVRG method uses randomized smoothing technique to smooth the component functions and exploits the finite average structure of the smooth part by extending the variance reduction technique of SVRG, which computes the full gradient periodically to modify the stochastic gradients in order to reduce their variance. We have given, to the best of our knowledge, the first variance reduction techniques for large-scale nonsmooth convex optimization. 

From convergence analysis, we prove that RS-SVRG method enjoys linear convergence with constant convergence rate. Besides, it enjoys much lower time complexity and gradient complexity than gradient methods-stochastic subgradient descent and full subgradient descent. In addition, compared with some state-of-art nonsmooth algorithms including Nesterov's smoothing and Bundle method, our method does not require the storage of the historical subgradient information on each component functions, which saves significant computational budge on large-scale and high dimensional problems. It should be noted that our method can be applied to a more general class of problems, without strongly-convex condition on objective and any other global and local error bound condition. 

We study the effects of different smoothing distributions on our algorithm, and derive several corollaries outlining upper convergence rate bounds with the problem dimension and number of smoothing samples. Our experiments also show qualitatively good agreement with the theoretical predictions we have developed. 

\makeatletter
\def\@biblabel#1{}
\makeatother
\bibliographystyle{apalike}
\bibliography{SVRG}

\end{document}